\documentclass[conference]{IEEEtran}
\IEEEoverridecommandlockouts
\usepackage{cite}
\usepackage{amsmath,amssymb,amsfonts}
\usepackage{algorithmic}
\usepackage{graphicx}
\usepackage{textcomp}
\usepackage{xcolor}
\def\BibTeX{{\rm B\kern-.05em{\sc i\kern-.025em b}\kern-.08em
    T\kern-.1667em\lower.7ex\hbox{E}\kern-.125emX}}

\begin{document}

\title{Multi-Subset Approach to Early Sepsis Prediction}

\author{\IEEEauthorblockN{Kevin Ewig\IEEEauthorrefmark{2}, Xiangwen Lin\IEEEauthorrefmark{2}, Tucker Stewart\IEEEauthorrefmark{2}, Katherine Stern\IEEEauthorrefmark{4}, Grant O'Keefe\IEEEauthorrefmark{4}, Ankur Teredesai\IEEEauthorrefmark{2,3}, and Juhua Hu*\IEEEauthorrefmark{2}}
\IEEEauthorblockA{\IEEEauthorrefmark{2}School of Engineering and Technology, University of Washington, Tacoma, WA 98402, USA}
\IEEEauthorblockA{\IEEEauthorrefmark{4}Department of Surgery, University of Washington, Seattle, WA 98195, USA}
\IEEEauthorblockA{\IEEEauthorrefmark{3}CueZen Inc., Seattle, WA 98101, USA}
\IEEEauthorblockA{\{ewigkl, xlin8, trstew, kstern, gokeefe, ankurt, juhuah\}@uw.edu}
}

\maketitle

\begin{abstract}
Sepsis is a life-threatening organ malfunction caused by the host's inability to fight infection, which can lead to death without proper and immediate treatment. Therefore, early diagnosis and medical treatment of sepsis in critically ill populations at high risk for sepsis and sepsis-associated mortality are vital to providing the patient with rapid therapy. Studies show that advancing sepsis detection by 6 hours leads to earlier administration of antibiotics, which is associated with improved mortality. However, clinical scores like Sequential Organ Failure Assessment (SOFA) are not applicable for early prediction, while machine learning algorithms can help capture the progressing pattern for early prediction. Therefore, we aim to develop a machine learning algorithm that predicts sepsis onset 6 hours before it is suspected clinically. Although some machine learning algorithms have been applied to sepsis prediction, many of them did not consider the fact that six hours is not a small gap. To overcome this big gap challenge, we explore a multi-subset approach in which the likelihood of sepsis occurring earlier than 6 hours is output from a previous subset and feed to the target subset as additional features. Moreover, we use the hourly sampled data like vital signs in an observation window to derive a temporal change trend to further assist, which however is often ignored by previous studies. Our empirical study shows that both the multi-subset approach to alleviating the 6-hour gap and the added temporal trend features can help improve the performance of sepsis-related early prediction.
\end{abstract}


\section{Introduction}

Sepsis is a major public health concern. It is a life-threatening disease caused by a host's failed response to an infection \cite{Singer2016-st}. The immune system of a sepsis patient becomes aggressive in its protection against infection in the body, which causes organ dysfunction and potential organ damage. Sepsis is still a common problem in modern medical settings, particularly Intensive Care Units (ICUs). According to a global survey conducted in 2018 \cite{sakr2018sepsis}, roughly 13.6\%  to 39.3\%  of patients admitted to ICU are impacted by sepsis. This share is 29.5\%  worldwide. Patients with sepsis also experience longer and more expensive hospital stays. For patients that survive, sepsis can cause an increased risk of permanent organ damage, and physical disability \cite{Jones2013}. 

Early treatment before the formation of sepsis in patients has been shown to improve the chances of successfully treating and preventing the disease \cite{Kim2018}. Early care can prevent 80\%  of sepsis-related deaths, and the chances of survival drop by 8\% every hour if action is not taken \cite{kumar2006duration}. In particular, studies have shown that treating sepsis 6 hours earlier before the onset significantly improves the patient's chance of recovering \cite{Gauer2013}. Therefore, this work focuses on the problem of early sepsis prediction, that is, 6 hours in advance. Several clinical scores such as SOFA have been developed to indicate the onset of sepsis with clinical-based data \cite{Balk2013-jb, Singer2016-st}. These measures are helpful when sepsis is onset already but have been limited for early prediction. For early sepsis onset prediction, predictions derived using supervised machine learning models such as Random Forest or Long Short-Term Memory (LSTM) models have vastly outperformed clinical scores \cite{Vollmer2019}. 

Because of this, there has been ongoing research in developing machine learning predictive models to detect the onset of sepsis early before it is suspected clinically. Some studies have used traditional machine learning models such as Logistic Regression and Random Forest \cite{MahmudLR2019, MahmudRF2020}. However, these studies did not capture the progressing temporal pattern that can be useful for early sepsis prediction. Recently, deep learning models have also been applied in sepsis prediction, such as Recurrent Neural Networks (RNNs) and Convolutional Neural Networks (CNNs) \cite{LiXinNg2019}. Compared with traditional machine learning methods, deep learning often provides better prediction performance. However, training deep models can be tedious to obtain the optimal subsets and parameters. In addition, none of them has considered the fact that 6 hours is not a small gap. Without the progressing information for the next 6 hours before onset, it is challenging to make an accurate prediction, which is explored in this work. 

First, it should be noted that training successful deep neural networks (DNNs) is expensive in terms of time, computational resources, and data. We aim to avoid using DNNs but still capture the progressing temporal change pattern for early sepsis prediction. We accomplish this by computing the hourly changes in feature values like heart rate within a given observation window. Doing this will allow us to consider the temporal changes in the feature even in a traditional machine learning model, where features are often considered independently. Second, we propose a multi-subset approach to close the 6-hour gap. The contributions of this paper can be summarized as follows.

\begin{itemize}
\item We propose to generate a series of delta values to capture the temporal change trend, which is incorporated into the feature set for training and prediction.
\item We develop a multi-subset approach in which the likelihood of sepsis arising earlier than 6 hours are generated from the previous subset and provided to the target subset as additional features to close the big 6-hour gap.
\item Based on the proposed method, we apply an economical machine learning algorithm, that is XGBoost \cite{chen2016xgboost}, to the trauma patients from the year 2012 to 2019 at ICUs of UW Harborview Center, and obtain an AUROC of 0.7906 for early sepsis prediction and that of 0.9199 for early septic shock prediction, better than a deep learning model due to limited amount of sepsis cases.
\end{itemize}

\section{Related Work}

Various machine learning models have been studied for sepsis prediction as follows. 

\subsection{Traditional Models}

Some of the research done in sepsis prediction uses simple traditional machine learning models. For example, Zabihiet et al. \cite{Zabihi2019} used a wrapper feature selection algorithm based on XGBoost to extract five different sets of features from clinical data to predict sepsis 6 hours before onset. Both valid and missing clinical data are used to derive the relevant attributes. Afterward, an ensemble model comprised of five XGBoost models is utilized to predict sepsis. In addition, Firoozabadi et al. \cite{Firoozabadi2019} created models to predict sepsis as part of the PhysioNet/CinC Challenge 2019. In this study, the authors took hourly samples of 40 features from each patient's data from three different ICUs. They processed the data to remove outliers, fill in missing data, and replace the remainder with the population average. They discovered similarly that an ensemble of bagged decision trees is effective for early sepsis prediction 6 hours before onset. A similar observation is shown by Fu et al. \cite{Fu2019}.

However, these models did not consider that the temporal change trend can be helpful for early sepsis onset prediction, and missing 6 hours of progressing information is challenging to make an accurate prediction. In this work, we address both by adding temporal change trend features and proposing a multi-subset approach. 

\subsection{Deep Learning Models}

Deep learning models have also been applied to address specific challenges of sepsis prediction. For example, Gilbertson et al. \cite{Gilbertson2019} used the Principal Component Analysis (PCA) to reduce the dimensionality of patient data before applying a simple RNN. The PCA method reduced the original 40 patient features to 10 main components. In addition to this, a fast Sequential Organ Failure Assessment (qSOFA) score generates one additional characteristic. These 11 characteristics are then loaded into a DNN classifier. However, the characteristics connected with each hour are examined independently. Later, Tsang et al. \cite{Tsang2021} applied LSTM which allows consideration of past time-step data embeddings within the prediction of the current time-step to capture the temporal pattern. Most recently, Shah et al. \cite{Shah2021} proposed a DNN model to predict sepsis 6 hours in advance. The DNN model is used because of its feature learning capacity and its property of approximating functions.

Better sepsis onset prediction performances were observed using these DNNs. However, none of them has considered addressing the 6-hour gap challenge, which is explored in this paper. Moreover, training DNNs is often time, resource, and data expensive, which is avoided in this work. 

\subsection{Temporal Change Trend}
In 2019 and 2021, Brekke et al. \cite{pmid30645637} and Mahta et al. \cite{pmid34583497} respectively proved that the temporal trend of vital signs and white blood cells could improve the performance of deterioration prediction by studying the retrospective cohort. In the machine learning area, Orphanou et al. \cite{pmid29555443} employed Temporal Association Rules (TAR) combined with Naïve Bayes classifiers for coronary heart disease diagnosis. They showed that higher levels of temporal abstraction improved the prediction performance when long sequences and distant events are critical. 

In the sepsis-related prediction area, Ghosh et al. \cite{pmid28011233} adapted temporal patterns resulting from the contrast pattern mining approach in conjunction with Coupled Hidden Markov Model (CHMM) to predict septic shock 30-60 minutes prior to onset. This paper demonstrated that this performance is significantly higher than applying SVM directly on data or using CHMM with continuous multivariate data. Nonetheless, the 30-60 minutes prediction window is too narrow to let the doctor react. 

\section{The Proposed Method}

We assume that the Electronic Health Records (EHRs) for a single patient can be represented as a \begin{math}t \times f\end{math} matrix \begin{math}X\end{math}. There are \begin{math}t\end{math} number of rows, where each row represents an hourly observation of a patient's features and where the $t$-th observation is the most recent. There is also  $f$ number of columns, where each column represents a physiological feature. Let \begin{math}x_{t,f}\end{math} denote the numerical entry in $X$ for time $t$ and feature $f$.

\subsection{Multi-Subset Approach}

We denote the ground truth\textsuperscript{\textsection} for a sepsis incident as $\hat{y}$ that has an indication of onset hour $t$. More than often, the hourly physiological patient data contain missing values, which are imputed and filled in.  In order to train the model to predict the onset of a sepsis incident $h$ hours into the future, for each patient, we first shift the values in $\hat{y}$ earlier by $h$ hours and call this $\hat{y_h}$. This step creates a new target feature that will occur in $h$ hours. The leftover values on the table with no ground truth values are dropped from the data. This step is illustrated in Table~\ref{table1} when we set $h=3$. From this example, we can observe that if we want to predict sepsis event onset in 3 hours, using observed patient data up to $t=1$, the onset prediction for $t=4$ should be 0, and using data up to $t=2$, the prediction of the onset of a sepsis incident for $t=5$ should be 1.

\begingroup\renewcommand\thefootnote{\textsection}
\footnotetext{The onset of a sepsis incident label has been given using the CDC’s adult sepsis surveillance criteria with prior modifications utilizing readily obtainable EMR data to improve specificity for the trauma population.  It is required that all of the following be present: 1) an order for a new IV or qualifying oral antibiotic, not administered within the previous 48 hours and excluding antibiotics used for surgical prophylaxis; 2) a body tissue culture was ordered within 48 hours of antibiotic initiation; 3) a qualifying antibiotic was sustained for at least 4 consecutive days, or until death or discharge; and 4) a 2-point increase in the maximum daily sequential organ failure assessment (SOFA) score occurred within 3 days before and 3 days after the qualifying culture. The criteria are restricted to hospital-acquired infections, which are defined as cultures obtained on or after the third hospital day.  Two subgroups were independently adjudicated before final sepsis assignments were made: culture-negative sepsis and patients meeting partial but not full criteria.}
\endgroup

\begin{table}
	\caption{In this example, the table on the left contains the ground truth for a sepsis incident at time $t$. In order to produce new ground truth values for an incident at time $t + h$, the $\hat{y}$ values are shifted by $h = 3$ hours. The leftover rows are dropped.}
    \centering
    \renewcommand{\thetable}{\arabic{table}}
    \begin{tabular}{|l|l|}
    \hline
    \begin{math}t\end{math} & \begin{math}\hat{y}\end{math}        \\ \hline
    1 & 0        \\ \hline
    2 & 0        \\ \hline
    3 & 0        \\ \hline
    4 & 0        \\ \hline
    5 & 1        \\ \hline
    6 & 1        \\ \hline    
    \end{tabular} 	
    \quad
	$\rightarrow$
	\quad
    \begin{tabular}{|l|l|}
    \hline
    \begin{math}t\end{math} & \begin{math}\hat{y_3}\end{math}        \\ \hline
    1 & 0        \\ \hline
    2 & 1        \\ \hline
    3 & 1        \\ \hline
    4 & -        \\ \hline
    5 & -        \\ \hline
    6 & -        \\ \hline    
    \end{tabular}	
    \quad
	$\rightarrow$
	\quad    
    \begin{tabular}{|l|l|}
    \hline
    \begin{math}t\end{math} & \begin{math}\hat{y_3}\end{math}        \\ \hline
    1 & 0        \\ \hline
    2 & 1        \\ \hline
    3 & 1        \\ \hline
    \end{tabular}	
    \vspace{2mm}
	\label{table1}  
\end{table}

After shifting the label for $h$ hours to predict the onset of an incident in $h$ hours, for a certain hour $t$, we can use only the $x_{t,f}$ values across all of the features, and the new ground truth label $\hat{y_h}$ to train a classification model for the onset of a sepsis incident that will be able to make a prediction on new patients and provide the probability of the event in $h$ hours as $y_h$. This is illustrated as an example in Table~\ref{table2}. Specifically, given the hard label of an incident, the trained model can provide the probability of the onset of such an event at each time stamp, which inspires our multi-subset approach. 

\begin{table}
	\caption{Probability of a sepsis incident in $h = 3$ hours. Features $x_1$, $x_2$, ..., $x_f$ are used to train a model with the ground truth label $\hat{y_h}$, and the model is then used to predict the probability of onset in $h$ hours as $y_h$.}   
    \centering
    \renewcommand{\thetable}{\arabic{table}}
    \begin{tabular}{|l|l|l|l|l|}
    \hline
    \begin{math}t\end{math} & \begin{math}x_1\end{math} & \begin{math}x_2\end{math} & ... & \begin{math}x_f\end{math}                  \\ \hline
    1          & 98  & 102 & ... & 30          \\ \hline
    2          & 95  & 107 & ... & 32          \\ \hline
    3          & 98  & 111 & ... & 35          \\ \hline
    4          & 100 & 109 & ... & 40          \\ \hline
    5          & 97  & 109 & ... & 41          \\ \hline
    \end{tabular}
    \quad
	+
	\quad 
    \begin{tabular}{|l|}
    \hline
    \begin{math}\hat{y_3}\end{math}        \\ \hline
    0        \\ \hline
    0        \\ \hline
    0        \\ \hline
    0        \\ \hline
    1        \\ \hline
    \end{tabular}	
    \quad
	$\rightarrow$
	\quad
    \begin{tabular}{|l|}
    \hline
    \begin{math} y_3 \end{math}           \\ \hline
    0.16          \\ \hline
    0.24          \\ \hline
    0.32          \\ \hline
    0.49          \\ \hline
    0.61          \\ \hline
    \end{tabular}  
    \vspace{2mm}

	\label{table2}
\end{table}

\begin{figure}
    \centering
    \includegraphics[width=.49\textwidth]{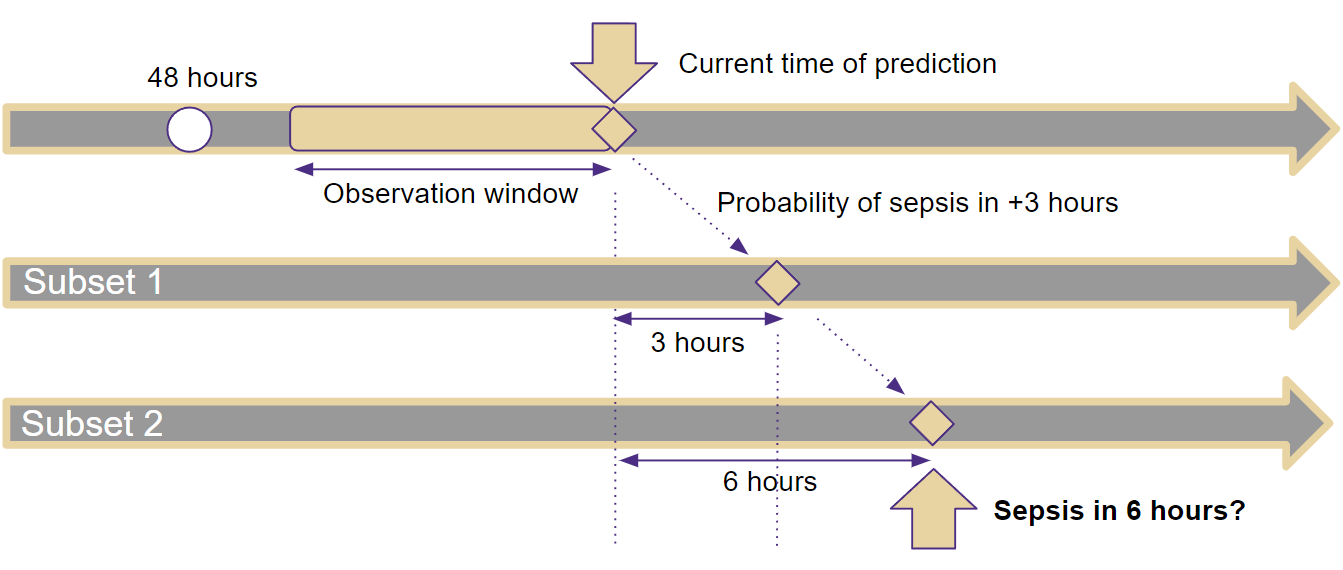}
    \caption{Timeline of subsets used to close the prediction gap}
    \label{fig1}
\end{figure}

To close the 6-hour gap of missing progressing information, we can include the probability of a sepsis incident in 3 hours as a new feature (i.e., one subset) to predict the incident in 6 hours in the target subset as shown in Table~\ref{table3}. This procedure is also illustrated in Fig.~\ref{fig1}. In a previous subset (e.g., Subset 1 in Fig.~\ref{fig1}), we use the observed data up to the prediction time to determine the probabilities that a patient will experience a sepsis incident between the current time $t$ and the sepsis incident onset time $t+6$. For example, we can use the observed data to determine the probability of the onset of a sepsis incident at $t+3$. Then, this probability will be treated as an additional input feature in the target subset (e.g., Subset 2 in Fig.~\ref{fig1}) of our approach. After that, the large 6-hour gap can be shrunk using an in-between measure with a smaller gap of 3. We can further shrink the gap by providing sepsis incident probabilities of each hour in-between to feed more features in the target subset, which is using multiple subsets to close the 6-hour gap.

\begin{table}
	\caption{A toy example of calculating the probability of onset in 6 hours with an additional feature of $y_3$, that is, the probability of onset at $h=3$ output from a previous subset.}  
    \centering
    \renewcommand{\thetable}{\arabic{table}}
    \begin{tabular}{|l|l|l|l|l|l|}
    \hline
    \begin{math}t\end{math} & \begin{math}x_1\end{math} & \begin{math}x_2\end{math} & ... & \begin{math}x_n\end{math} & \begin{math} y_3 \end{math}            \\ \hline
    1          & 98  & 102 & ... & 30  & 0.16        \\ \hline
    2          & 95  & 107 & ... & 32  & 0.24        \\ \hline
    \end{tabular}
    \quad
	+
	\quad 
    \begin{tabular}{|l|}
    \hline
    \begin{math}\hat{y_6}\end{math}        \\ \hline
    0        \\ \hline
    1        \\ \hline
    \end{tabular}	
    \quad
	$\rightarrow$
	\quad
    \begin{tabular}{|l|}
    \hline
    \begin{math} y_6 \end{math}           \\ \hline
    0.46          \\ \hline
    0.64          \\ \hline
    \end{tabular}  
    \vspace{2mm}
  
	\label{table3}
\end{table}

\subsection{Delta Values for Temporal Change Trend}
It should be noted that the same absolute value for different patients can have totally different meanings, while changes over time can share the same indication. Therefore, we use delta values to capture the changes in feature values over time, which can provide valuable temporal patterns for sepsis prediction. The observation window size as shown in Fig.~\ref{fig1}, denoted by $w$, is an important parameter that determines the duration of time over which the delta values are computed.

To compute the delta values, we create a new feature, denoted by $f'$, which is computed as the difference between the value of $f$ at timestamp $t$ and the value of $f$ at the previous timestamp $t-1$ within the observation window $w$. In other words, the value of $f'$ at timestamp $t$ is calculated as $x_{t,f'} = x_{t,f} - x_{t-1,f}$. This computation is repeated for each feature and each timestamp within the observation window. For example, if the heart rate at timestamp 11 is 70 beats per minute (bpm) and the heart rate at timestamp 10 is 75 bpm, the delta value for the heart rate at timestamp 11 would be -5 bpm.

\subsection{Statistical Values for Temporal Change Trend}

We leverage delta values to capture the temporal change trend within the observation window $w$. We can also incorporate statistical information computed from a feature $f$ within the same time window to describe the feature distribution. Specifically, for each feature $f$ within the observation window $w$, we can calculate six statistical values based on the sequence $S = [x_{t-w,f}, x_{t-w+1,f}, x_{t-w+2,f}, ..., x_{t,f}]$: the mean, minimum, maximum, standard deviation, skewness, and kurtosis. Variance, which is the expectation of the squared deviation, measures how spread out the data is, while skewness is a measure of the asymmetry of the variable's probability distribution, and kurtosis measures the tailedness of the probability distribution. The formulas for calculating these statistical values are as follows: 

\begin{itemize}
\item Mean value: $ \overline{x}= \frac{1}{N} \sum_{i=1}^{N}x_i $ 
\item Standard deviation: $ \sigma =  \sqrt{ \frac{1}{N} \sum_{n=1}^{N}(x_i - \overline{x})^2 } $
\item Kurtosis: $\text{K} = \frac{1}{N} \sum_{i=1}^{N} \frac{(x_i -\overline{x})^4 }{ \sigma ^4} $ 
\item Skewness: $\text{Sk} = \frac{1}{N} \sum_{i=1}^{N} \frac{(x_i- \overline{x})^3 }{ \sigma ^3} $
\end{itemize}

\begin{figure}
    \centering
    \includegraphics[width=0.4\textwidth, height=0.5\textheight]{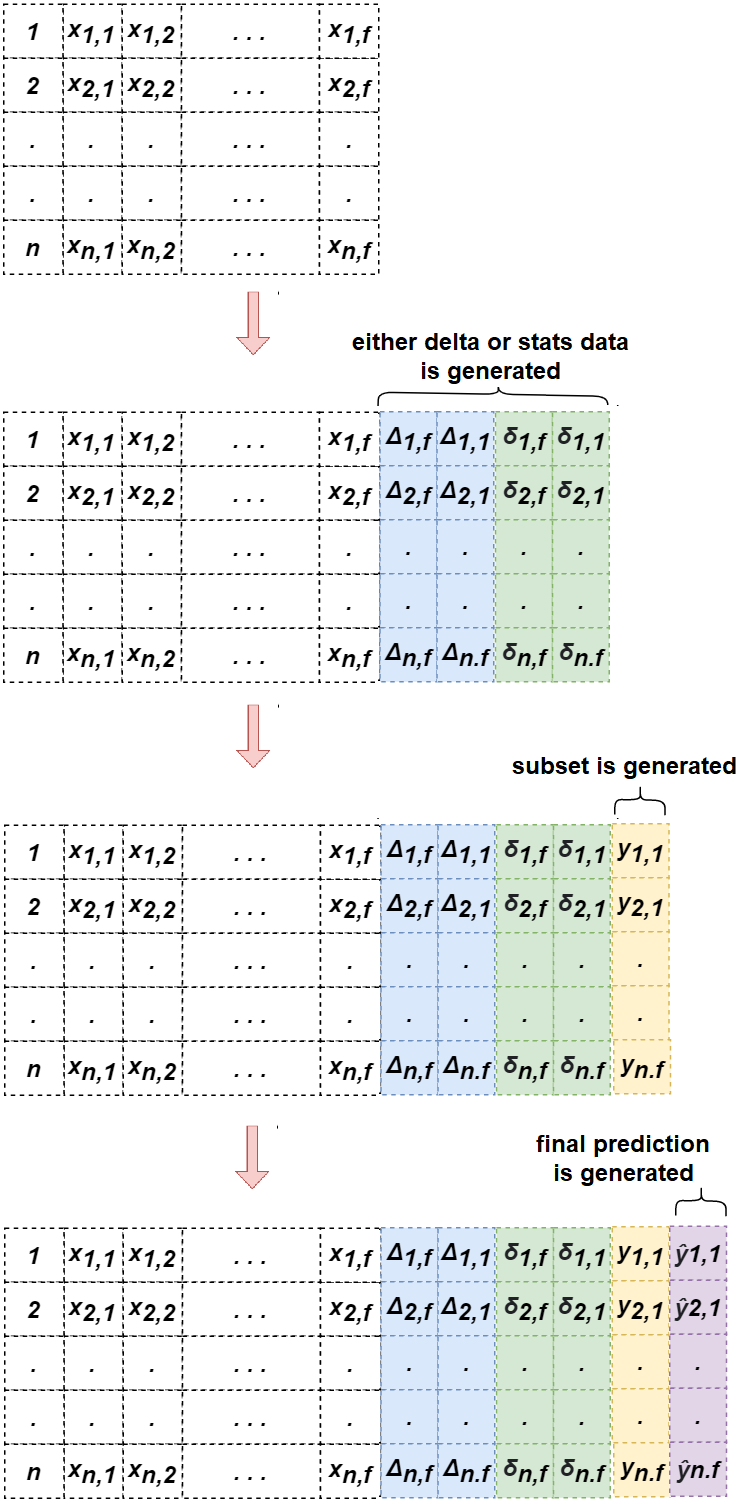}
    \caption{An overview of our proposed method.}
    \label{fig2}
\end{figure}

By incorporating both delta values and statistical information, our model is better equipped to capture the temporal change trend for early sepsis prediction. The overall process is illustrated in  Fig.~\ref{fig2}. First, delta and/or statistic values are generated. Then, a 3-hour prediction is generated to address the prediction gap and added as a previous subset for the target 6-hour prediction.

\section{Experiments}

To sufficiently demonstrate the proposed method, we apply it to two important sepsis-related tasks, that is, early sepsis onset prediction and early septic shock prediction. we conduct extensive experiments using the state-of-art traditional machine learning model, i.e., XGBoost \cite{chen2016xgboost}, to justify the superiority of our proposal and answer the following questions:

\begin{itemize}
\item What are the benefits of using multi-subsets in the development of sepsis and septic shock prediction models?
\item How do delta and statistic values incorporate the temporal change trend affect the effectiveness of the prediction models for sepsis and septic shock?
\item Which features are more important for early sepsis or septic shock prediction?
\end{itemize}

\subsection{Data Description}
To evaluate the proposed method, we use trauma patients' data from UW Harborview Medical Center. It contains prospectively collected, de-identified data from injured adults ages 16 years and older, who were admitted to the ICU between 2012 to 2019 and required at least three days of invasive mechanical ventilation. This data includes physiological data on 2,802 patients, 486 (17\%)  were identified to have sepsis during the first 14 days of admission. The patient data is sampled at different time intervals. Vital signs, for example, were sampled hourly, whereas laboratory tests were sampled daily or less frequently. As such, there are some features with many missing values. We address missing values using carry-forward imputation. 

To find which features would be more useful for early sepsis prediction, we set groups of features that share common attributes. They are vital signs, static profile, cumulative exposures, and laboratory results, where static profile includes initial physiology from the first 48 hours and patient factors, and cumulative exposures are the summary of events. These 4 groups are denoted as G1, G2, G3, and G4. The detailed list is shown in the Appendix.

\subsection{Imbalanced-Class Problem in Data}

One of the primary problems with the sepsis patient data we received is predicting an outcome that occurs in a minority of the study population using an hour-by-hour framework. This can be a problem because it leads to imbalanced classes in the data. In this case, the outcome of interest, which is a sepsis event, occurs in only 17\% of the patients, making it a minority class. When there is a class imbalance in the data, it can lead to biased model performance. Approximately 1.84\% of the hour-by-hour records in the data have a sepsis indicator, while the rest does not have it. Most of the patients admitted to the ICU never develop sepsis. Some patients develop sepsis within an hour or two, and some develop sepsis after a more extended period.   
 
To account for the imbalanced nature of the data, we use the following methods. First, we only include hourly patient records from day 2 until day 14 of the patient's stay in the ICU. This is because the scope of this study is patients developing sepsis after being in the ICU. Limiting the data will reduce the number of hourly records with no sepsis that does not need to be included. Second, if a patient has an hourly record with sepsis, we remove all the subsequent hourly records for that patient. For example, if we see a patient record with sepsis at 4:00, all observations after that time are removed. It is because we are only interested in the early sepsis onset prediction, that is, the first onset. In addition, to further account for the imbalance of data, we use a random oversampling of 0.8. To illustrate this further, suppose there are 20 minority samples and 1000 majority samples. Random oversampling duplicates the 20 minority samples without replacing them until there are 800 samples. As a result, there are 800 minority samples and 1000 majority samples. Finally, we use random under-sampling to make the number of minority class data the same as the number of majority class data in the sample. 

\subsection{Evaluation Metrics}

To evaluate the performance of the proposed approach, we use metrics of Area Under the Receiver Operator Curve (AUROC), Sensitivity, and Specificity, where the average performance under 5-fold cross-validation is reported. Specifically, \begin{math}Sensitivity = \frac{TP}{TP+FN}\end{math}, where $TP$ is the number of true positives and $FN$ is the number of false negatives, is defined as a measure of how well the model can recognize positive examples. 

At the same time, \begin{math}Specificity = \frac{TN}{FP+TN}\end{math} is the fraction of real negatives properly detected by the model. Finally, AUROC is a useful metric for evaluating the performance of a binary classifier because it provides a single number that summarizes the classifier's ability to discriminate between positive and negative samples over all possible decision thresholds.


\subsection{Effect of Multi-Subset Approach}

Considering the feature selection property of XGBoost that can benefit from more features, we use the largest set of features (i.e., Group 1+2+3+4 with delta and statistic values in the observation window, which is set to $w=6$) to demonstrate the effect of our multi-subset approach. In the comparison, we use `1 Subset' to denote the method using only the target subset, `2 Subsets' to denote the method adding the probability of sepsis in 3 hours to the target subset, and `6 Subsets' to denote the method adding probabilities of sepsis in 1 hour, 2 hours, 3 hours, 4 hours, and 5 hours to the target subset.

\begin{table}[h]
\caption{Performance Comparison of Multi-subsets with Group 1+2+3+4 delta and stats for Sepsis and Septic shock prediction.}
\begin{center}
\begin{tabular}{|l|l|l|l|}
\hline
\textit{\textbf{Name}} & \textit{\textbf{AUROC}} & \textit{\textbf{Sensitivity}} & \textit{\textbf{Specificity}} \\ \hline
\multicolumn{4}{|c|}{Sepsis}\\\hline
1 Subset	 & 	0.7720 & 	\textbf{0.7090} & 	0.6960 \\ \hline
2 Subsets    & 	\textbf{0.7790} & 	0.6990 & 	\textbf{0.7220} \\ \hline
6 Subsets  &     0.6180 & 	0.6210 & 	0.5490 \\ \hline 
\multicolumn{4}{|c|}{Septic Shock}\\\hline
1 Subset	        & 0.7906           & 0.6000               & 0.6188 \\ \hline
2 Subsets           & 0.8219           & 0.6971            & 0.7587 \\ \hline
6 Subsets  & \textbf{0.8781}  & \textbf{0.7943}   & \textbf{0.7829} \\ \hline
\end{tabular}
\label{table-benefitofmultisubsets-3}
\end{center}
\end{table}

Table~\ref{table-benefitofmultisubsets-3} compares the performance for both the tasks of early sepsis prediction and early septic shock prediction. We can observe that our multi-subset approach is definitely helpful for the task of early septic shock prediction, where adding the probability of shock in 3 hours improves the performance compared to using only the target subset and adding more probabilities can further improve. However, this is not the case for early sepsis prediction. In the experiment, we observed that more features does not mean more choices of features for XGBoost that can help improve the performance. For the task of early sepsis prediction, we find that using only Group 1 (i.e., vital signs) and Group 2 (i.e., static profile) features are more helpful as shown in Table~\ref{table-benefitofmultisubsets-1}. More importantly, we can observe the similar phenomenon that adding the probability of sepsis in 3 hours improves the performance compared to using only the target subset and adding more probabilities can further improve.

\begin{table}[h]
\caption{Performance Comparison of Multi-subsets with Group 1+2 delta and stats for Sepsis prediction.}
\begin{center}
\begin{tabular}{|l|l|l|l|}
\hline
\textit{\textbf{Name}} & \textit{\textbf{AUROC}} & \textit{\textbf{Sensitivity}} & \textit{\textbf{Specificity}} \\ \hline
1 Subset                 & 0.6072           & 0.5741           & 0.5917  \\ \hline
2 Subsets                & 0.7111           & 0.6420            & 0.6695  \\ \hline   
6 Subsets       & \textbf{0.7906}  & \textbf{0.7222}  & \textbf{0.7251}  \\ \hline
\end{tabular}
\label{table-benefitofmultisubsets-1}
\end{center}
\end{table}

\begin{figure}[h]
    \centering
    \includegraphics[width=.48\textwidth]{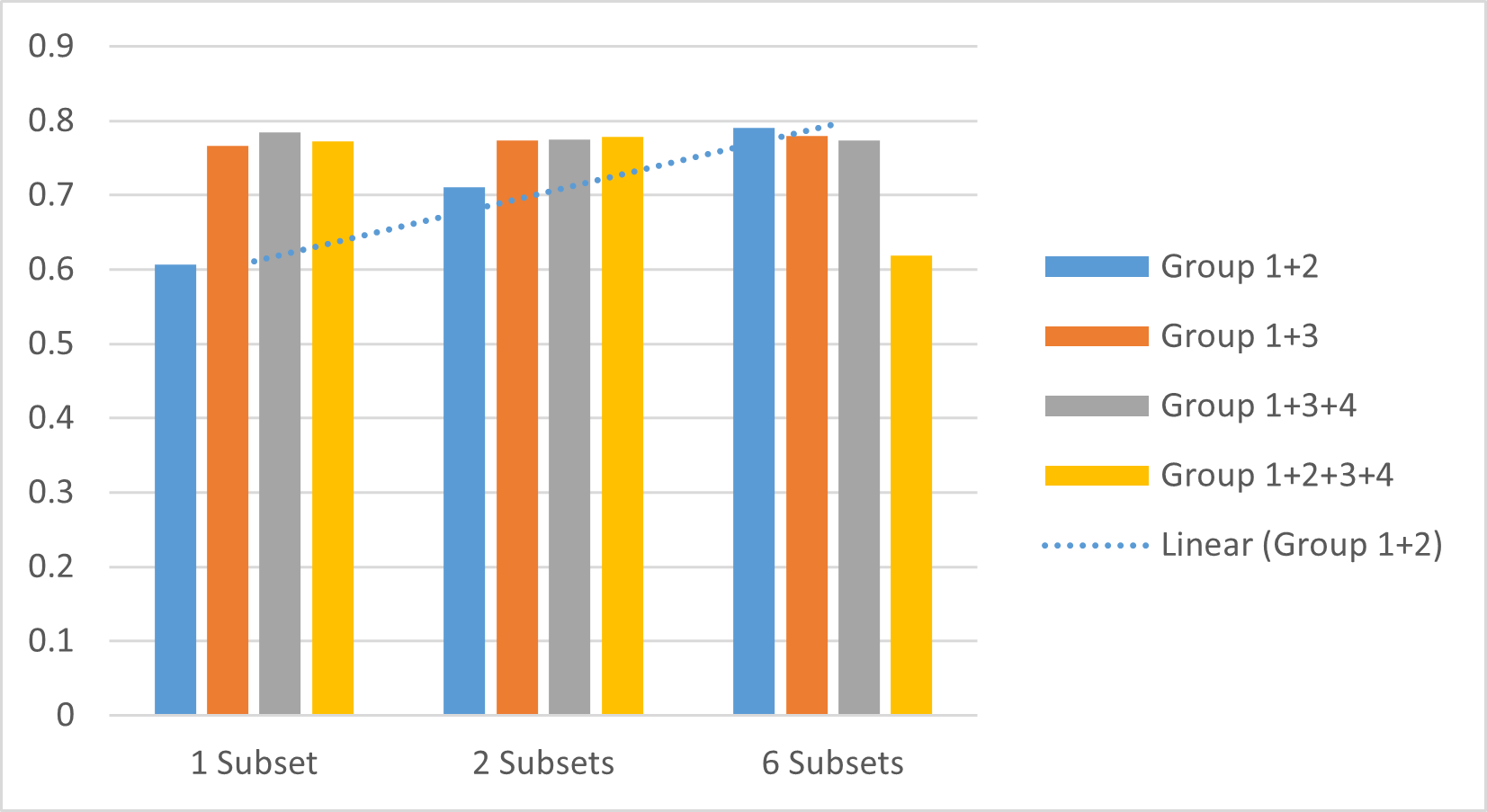}
    \caption{AUROC comparison over different groups with delta \& stats and different subsets for sepsis onset prediction.}
    \label{fig3}
\end{figure}

Fig.~\ref{fig3} further demonstrates the observation that for early sepsis prediction, certain features are not helpful and even harmful. In general, the model with more subsets performs better than the model that has only 1 subset. However, we find that the group combinations with G3 and G4 may affect the result lower than fewer subsets. For example, Fig.~\ref{fig3} shows the result of different group combinations with delta and stats using different subsets. Group 1+2 and Group 1+3 are getting better when adding subsets, but Group 1+3+4 is getting lower, and Group 1+2+3+4 is getting higher first, then getting lower. This is because that G4 has much more missing values other groups, which may mislead the model learning.

\subsection{Effect of Delta and Stats} 

In this sub-section, we evaluate the effect of using delta, stats, or a combination of delta and stats in the prediction for sepsis and septic shock. We take the best feature group combination for each task with `6 Subsets' to compare, that is, Group 1+2 for sepsis and Group 1+2+3+4 for septic shock, respectively. 

\begin{table}[h]
\caption{Performance Comparison with Group 1+2 for Sepsis and Group 1+2+3+4 for Septic Shock using `6 Subsets'.}
\begin{center}
\begin{tabular}{|l|l|l|l|}
\hline
\textit{\textbf{Name}} & \textit{\textbf{AUROC}} & \textit{\textbf{Sensitivity}} & \textit{\textbf{Specificity}} \\ \hline
\multicolumn{4}{|c|}{Sepsis}\\\hline
Baseline        & 0.7495 &	0.6667 &	\textbf{0.7277} \\ \hline
Delta           & 0.7629 &	0.6728 &	0.7192  \\ \hline
Stats  & 0.7904 &	0.7119 &	0.7170 \\ \hline   
Delta and Stats & \textbf{0.7906} &	\textbf{0.7222} &	0.7251 \\ \hline
DNN & 0.6869 & 0.4845 & 0.7824\\ \hline
\multicolumn{4}{|c|}{Septic Shock}\\\hline
Baseline        & \textbf{0.9199}  & \textbf{0.8457}     & \textbf{0.8037} \\ \hline
Delta           & 0.7929           & 0.7486             & 0.7021  \\ \hline
Stats           & 0.7943           & 0.7314             & 0.7319  \\ \hline   
Delta and Stats & 0.8781           & 0.7943	            & 0.7829  \\ \hline
DNN & 0.6098 & 0.3143 & 0.7734 \\ \hline
\end{tabular}
\label{table-benefitofdelta-1}
\end{center}
\end{table}

Table~\ref{table-benefitofdelta-1} shows the results. First, for sepsis prediction, we can observe that both AUROC and Sensitivity are increasing when incorporating delta and stats. This indicates that using delta and stats can help improve the detection performance of sepsis patients, while the baseline can be biased to non-sepsis patients with a way higher Specificity compared to the Sensitivity. This demonstrates that using temporal change trend is helpful for early sepsis prediction. 

For the task of septic shock, we have a different observation. Although the combination of Delta and Stats provides better performance compared to using only one of them, the baseline without using any additional temporal trend features gives the best performance. In fact, this is a reasonable observation, since shock as a sudden event is more related to the most recent conditions of the patient, while sepsis can be more observable using the progressing status of infections. In this case, using multiple subsets to shrink the 6-hour gap is more effective.

Moreover, considering the out-performance of DNNs, we apply RNN to the temporal features and DNN to static features. Then, we concatenate them for sepsis and septic prediction using the same setup. However, DNN performs worse. The potential reason is that DNNs often require sufficient training data to be successful, while we have very limited amount of sepsis cases, resulting in models significantly biased to Specificity.

\subsection{Model Explanation}

To further verify if these probabilities added from the previous subsets are helpful for making predictions, we apply LIME \cite{lime}, a tool that attempts to explain the weights in a predictive model, and check the importance of features for sepsis onset prediction. Specifically, we take a random sample of the data and apply LIME to get the importance of each feature for both models of `2 Subsets' and `6 Subsets' shown in Table~\ref{table-lime}. The top-5 important features of each model are listed.

\begin{table}[h]
    \caption{Top features from LIME}
    \begin{center}
    \renewcommand{\thetable}{\arabic{table}}
    \begin{tabular}{|l|r|}
        \multicolumn{2}{c}{3 Hour Probability Included}\\
        \hline
        Feature  & Value  \\
        \hline          \textbf{sepsis-subset-3hr}   & \textbf{0.026}  \\
        \hline          creatinine-6        & $-0.021$ \\
        \hline          surgSum-8           & $-0.020$ \\
        \hline          neutrophils-1       & $0.020$  \\
        \hline          neutrophils-9       & $0.019$  \\      
        \hline
    \end{tabular}
    \quad
    \begin{tabular}{|l|r|}
        
        \multicolumn{2}{c}{1-5 Hour Probabilities Included}\\
        \hline
        Feature  & Value  \\
        \hline          creatinine-3         & $-0.023$  \\
        \hline          neutrophils-2        & $-0.021$  \\
        \hline          neutrophils-5        & $0.019$   \\
        \hline          \textbf{sepsis-subset-5hr}    & \textbf{-0.019}  \\
        \hline          creatinine-6         & $0.014$   \\      
        \hline
    \end{tabular} 
    \end{center}
    \label{table-lime}
\end{table}

The results show that the probability of sepsis occurring in 3 hours contributed significantly to the prediction of sepsis onset for the method using 2 subsets. Similarly, the likelihood of sepsis occurring in 5 hours contributed significantly to the prediction for the method using 6 subsets. Moreover, the rest top features are delta values. This further demonstrates our proposal of shrinking the gap and using the temporal change trend.

\section{Conclusion}

In this work, we present an economical machine learning approach for sepsis related early prediction, which is way less expensive in terms of data, resources, and training time compared to deep learning models. First, to do early prediction in 6 hours, we propose a multi-subset approach to close the gap of missing 6 hours of patient data. Moreover, to capture the temporal change trend that can share similar meaning for different patients, we construct a series of delta and statistic values. Our extensive experiments demonstrate the effectiveness of these two components considering different scenarios, which is also shown in the model explanation. In the future, we would like to enhance the explainability of the model so that it can provide more valuable data, which in turn can identify discriminatory features to monitor and screen for sepsis related events.

\section{Acknowledgment}
Stewart and Hu’s research is supported in part by
NSF (IIS-2104270). Teredesai’s research is supported in part by
CueZen Inc. We would like to thank Microsoft Inc.,and University of Washington eScience Institute- Azure Cloud Computing Credits for Research and Teaching grant for their support by providing us with the Cloud Compute Resources on Microsoft Azure. All opinions, findings, conclusions and recommendations in this paper are those of the
authors and do not necessarily reflect the views of the funding
agencies.

\bibliographystyle{IEEEtran}
\bibliography{ref}

\section{Appendix}

\subsection{Feature Groups}
\footnote{* means this feature is used for generating delta and stats values}
G1 (Vital Signs)*:
\begin{itemize}
\item Heart Rate (hr) 
\item Diastolic Blood Pressure (dbp)  
\item Mean Arterial Pressure (map) 
\item Respiratory Rate (rr) 
\item Temperature (temp) 
\item FIO2 (fio2) 
\end{itemize}

G2 (Static Profile):
\begin{itemize}
\item Age 
\item Sex (male or female) 
\item Transfer from another hospital (transfer) 
\item Mechanism (Injury Mechanism) 
\item Head Injury 
\item ED Systolic Blood Pressure (Initial.ED.SBPCat)
\item Reverse Shock Index (rSICat) 
\item Max Base Deficit Cat (baseDef48Cat) 
\item Lactate Category (lactate48Cat) 
\item Red blood cell count (rbc48) 
\item Crystalloids count (crys48Cat) 
\item Apache 
\item Antibiotics 48 (abx48) 
\item Surgery 48 (surg48) 
\item ER Disposition (er dispCat) 
\end{itemize}

G3 (Cumulative Exposures):
\begin{itemize}
\item IV Fluid BOLUS (bolusSum) *
\item Red Blood Cell Count (RBCsum) 
\item Days in Ventilator (ventDaySum) 
\item Days in Surgery (surgSum) 
\item Hours in Surgery (surgHours)  
\end{itemize}

G4 (Laboratory Result):
\begin{itemize}
\item Bicarbonate (bicarb) *
\item Acidosis (acidosisCat)  
\item Strong Ion (StrongIon) *
\item Blood Urea Nitrogen (bun) *
\item Creatinine (creatinine) *
\item White blood cell count (wbc) 
\item Urine Output  (uop) 
\end{itemize}

\end{document}